\def\doublecolumn{1} 
\if 1\doublecolumn
  \documentclass[journal]{IEEEtran}
\else
  \documentclass[12pt, draftclsnofoot, onecolumn, letterpaper]{IEEEtran}
\fi

\def\blind{1} 

\usepackage[latin9]{inputenc}
\usepackage{algorithm, algorithmic}

\usepackage{array}

\usepackage{makecell} 
\usepackage{float}
\usepackage{mathtools}
\usepackage{amsmath}
\usepackage{amsthm}
\usepackage{amssymb}
\usepackage{graphicx}
\usepackage{wasysym}
\usepackage{setspace}
\usepackage{color}
\usepackage{bm}
\usepackage{cases}
\usepackage{tikz}
\usetikzlibrary{calc}
\usepackage{flowchart}
\usepackage{bbm}
\usepackage{adjustbox}
\usepackage{hhline}
\usepackage{makecell}

\usepackage{epsfig}
\usepackage{cite}
\usepackage{graphicx}
\usepackage{lipsum}
\usepackage{multirow}
\usepackage{array}
\usepackage{enumerate}
\usepackage{enumitem}
\usepackage{graphicx}
\usepackage{times}
\usepackage[framemethod=tikz]{mdframed}
\definecolor{cccolor}{rgb}{1,1,1}

\usepackage{subfig}
\usepackage{caption}
\newtheorem{remark}{Remark}

\usepackage{multicol,multirow,adjustbox}
\usepackage{booktabs}

\newcommand{\ie}[0]{\textit{i.e.}}

\usepackage[capitalize]{cleveref}
\crefname{section}{Sec.}{Secs.}
\Crefname{section}{Section}{Sections}
\crefname{table}{Tab.}{Tabs.}
\Crefname{table}{Table}{Tables}

\makeatletter
\def\@IEEEsectpunct{.\ \,}
\def\paragraph{\@startsection{paragraph}{4}{\z@}{1.5ex plus 1.5ex minus 0.5ex}%
{0ex}{\normalfont\normalsize\bfseries}}
\makeatother

\usepackage{pifont,xspace} 
\newcommand{\cmark}{\ding{51}\xspace}%
\newcommand{\cmarkg}{\textcolor{lightgray}{\ding{51}}\xspace}%
\newcommand{\xmark}{\ding{55}\xspace}%
\newcommand{\xmarkg}{\textcolor{lightgray}{\ding{55}}\xspace}%
\newcommand{\tmarkg}{\textcolor{lightgray}{\ding{115}}\xspace}%

\begin{document}

\title{
    Replace-then-Perturb: Targeted Adversarial Attacks With Visual Reasoning for Vision-Language Models 
}

\if 1\blind
\author{
    Jonggyu Jang,~\IEEEmembership{Member,~IEEE}, 
    Hyeonsu Lyu,~\IEEEmembership{Student Member,~IEEE}, 
    Jungyeon Koh, 
    and  Hyun~Jong~Yang,~\IEEEmembership{Member,~IEEE}
    \thanks{
    J. Jang is with Institute of New Media and Communications, Seoul National University, Seoul 08826, Republic of Korea, (email: jgjang@snu.ac.kr).
    H. Lyu and J. Koh are with Department of Electrical Engineering, Pohang University of Science and Technology (POSTECH), Pohang 37673, Republic of Korea, (e-mail: \{hslyu4, jwe06020\}@postech.ac.kr). 
    H. J. Yang (corresponding author) is with Department of Electrical and Computer Engineering, Seoul National University, Seoul 08826, Republic of Korea, (email: hjyang@snu.ac.kr).
    }
}
\else
\author{Anonymous Submission
    }
\fi

\maketitle

\begin{abstract}
    The conventional targeted adversarial attacks add a small perturbation to an image to make neural network models estimate the image as a predefined target class, even if it is not the correct target class. 
    Recently, for \textit{visual-language models} (VLMs), the focus of targeted adversarial attacks is to generate a perturbation that makes VLMs answer intended target text outputs. 
    For example, they aim to make a small perturbation on an image to make VLMs' answers change from ``\textit{there is an apple}'' to ``\textit{there is a baseball}.''
    However, answering just intended text outputs is insufficient for tricky questions like ``\textit{if there is a baseball, tell me what is below it}.''
    This is because the target of the adversarial attacks does not consider the \textit{overall integrity} of the original image, thereby leading to a lack of visual reasoning.
    In this work, we focus on generating targeted adversarial examples with \textit{visual reasoning} against VLMs. 
    To this end, we propose 1) a novel adversarial attack procedure---namely, \texttt{Replace-then-Perturb} and 2) a contrastive learning-based adversarial loss---namely, \texttt{Contrastive-Adv}.
    In \texttt{Replace-then-Perturb}, we first leverage a text-guided segmentation model to find the target object in the image. Then, we get rid of the target object and inpaint the empty space with the desired prompt.
    By doing this, we can generate a target image corresponding to the desired prompt, while maintaining the \textit{overall integrity} of the original image.
    Furthermore, in \texttt{Contrastive-Adv}, we design a novel loss function to obtain better adversarial examples.
    Our extensive benchmark results demonstrate that \texttt{Replace-then-Perturb} and \texttt{Contrastive-Adv} outperform the baseline adversarial attack algorithms.
    \textit{We note that the source code to reproduce the results will be available. }
\end{abstract}

\section{Introduction\label{sec:intro}}

\IEEEPARstart{R}{ecently,} vision-language models (VLMs) such as GPT-4~\cite{achiam2023gpt}, Claude 3.5~\cite{claude2024}, LLAVA 1.6~\cite{liu2024visual}, and Gemini~\cite{team2023gemini} have garnered significant research attention. 
The key abilities of VLMs are their capacity to understand and generate human-like responses to multi-modal inputs, thereby opening up new possibilities in numerous applications. 
For example, VLMs excel at processing and integrating both visual and textual information, enabling image captioning, visual question answering, and cross-modal retrieval. 
However, as VLMs continue to advance, their potential impact on both industry and academia is immense, thereby driving the need for further exploration into their capabilities and vulnerabilities.

\begin{figure}
    \centering
    \includegraphics[width=0.95\linewidth]{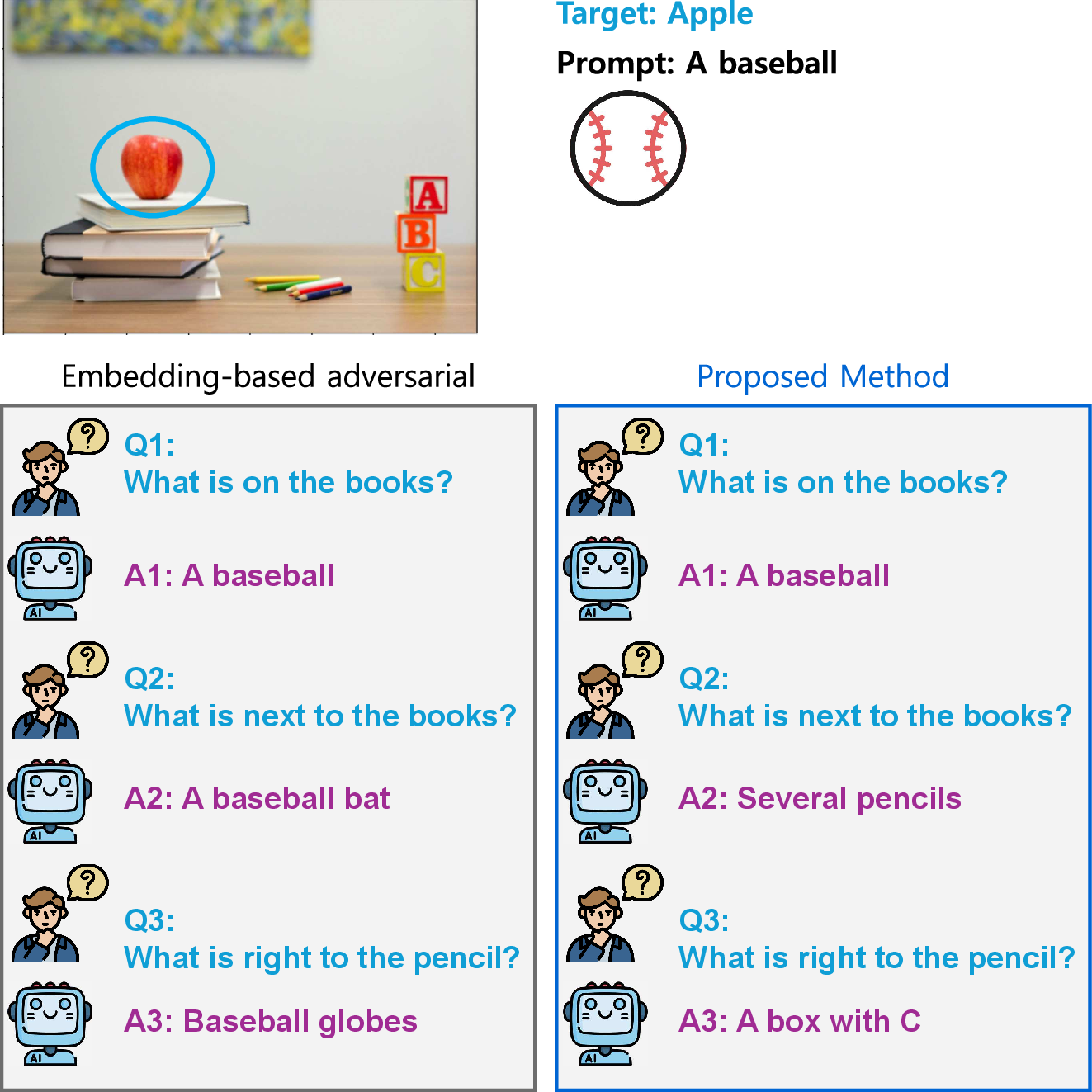}
    \caption{
    An example comparing embedding-based adversarial attacks and the proposed method. 
    The original image depicts books, alphabet blocks, pencils, a drawing, and an apple. 
    The target object is the apple in the image, and the desired prompt is ``a baseball." 
    (Left) In embedding-based adversarial attacks, the image is recognized as a baseball; however, due to a lack of visual reasoning, the VLMs provide unnatural outputs (\textbf{Q2, Q3}). 
    (Right) In the proposed method, incorporating visual reasoning, the VLMs generate natural outputs, correctly replacing the target object (\textbf{apple}) with a (\textbf{baseball}).
    }
    \label{fig:motivation_vs_baselines}
\end{figure}

\begin{table*}[!ht]
    \caption{ Summary of the existing studies on adversarial attacks against VLMs. }
    \centering
    \adjustbox{width=1\linewidth}{
    \begin{tabular}{ccccccccc}
    \toprule
        Ref. \# &  Application & Attack Type & Targeted & Visual Perturbation   & Generalization & Visual Reasoning & \makecell{Shared Vision-Language \\  Semantic Space} & Pub. year \\
    \midrule
        \cite{lu2024test} & VQA & Backdoor & \cmark & \cmark & \xmarkg & \xmarkg &  - &2024 \\
        \cite{zhang2022towards} & VQA & Adversarial & \xmarkg & \cmark & \cmark & \xmarkg & - & 2022 \\
        \cite{xu2018fooling} & VQA & Adversarial & \cmark & \cmark & \xmarkg & \xmarkg & - & 2018 \\
        \cite{luo2024an} & VQA & Adversarial & \cmark & \xmarkg & \tmarkg & \xmarkg & \cmarkg & 2024 \\
        \cite{lu2023set} &  Image Captioning & Adversarial & \cmark & \cmark & \xmarkg  & \xmarkg & \cmarkg & 2023 \\ 
        \cite{zhou2023advclip} & image-text retrieval & Backdoor & \cmark & \cmark & \xmarkg & \xmarkg & \cmarkg & 2023 \\ 
        \cite{zhao2023on} & VQA & Adversarial & \cmark & \cmark & \xmarkg & \xmarkg & \cmarkg & 2024 \\ 
        \cite{ye2024mutual} & VQA & Adversarial & \cmark & \cmark & \xmarkg & \xmarkg & \cmarkg & 2024 \\ 
        \midrule
        \textbf{Ours} & VQA / Image Captioning & Adversarial & \cmark & \cmark & \cmark & \cmark & \xmark & - \\
    \bottomrule
    \end{tabular}
    }
    \label{tab:existing}
\end{table*}

\paragraph*{Motivation} 
Driven by research interests in the vulnerabilities of VLMs, a key research direction is generating \textit{adversarial examples} (\ie, adversarial attacks), as they are closely related to several security and privacy applications such as adversarial training~\cite{carlini2022certified, cohen2019certified}, privacy protection~\cite{jang2024unveiling, su2023hiding}, and steganography~\cite{baluja2017hiding}.
Recently, several studies have focused on \textit{targeted} and \textit{type-II}\footnote{Adversarial attacks aim to deceive neural network models either by producing the same output with a significant change in the input (type-I) or by producing different outputs with negligibly small changes in the input (type-II)~\cite{lu2017safetynet}.} adversarial attacks~\cite{ye2024mutual, luo2024an, lu2023set, zhou2023advclip}. 
However, as depicted in \cref{fig:motivation_vs_baselines}, previous embedding-based methods lack visual reasoning, thereby providing unnatural responses to questions. 
In this study, we aim to answer the following research question: ``\textit{\textbf{How can we preserve the overall integrity of adversarial examples against VLMs?}}''.

\subsection{Existing Studies and Challenges}

\paragraph*{Existing works}

Several previous studies have proposed methods for generating adversarial examples against VLMs~\cite{gu2024a, lu2024test}. 
In \cite{lu2024test}, the authors propose a targeted adversarial attack algorithm aimed at forcing VLMs to produce predefined answers (e.g., ``I don't know'') when the input prompt includes a trigger pattern. 
Similarly, the authors of~\cite{xu2018fooling} have developed methods for generating adversarial examples against Visual Question Answering (VQA) models. 
In~\cite{zhang2022towards}, the authors introduce a \textit{non-targeted} adversarial attack on VLMs, aiming to cause VLMs to fail in providing proper answers.


\paragraph*{Challenge 1---Visual Reasoning}

Unlike classifiers, \textit{targeted} adversarial attacks on VLMs need to preserve the outputs for untargeted objects, whether or not adversarial perturbations are applied. For instance, in classifiers trained on ImageNet, it is sufficient for attackers to change the output from ``cat'' to ``dog.'' However, for VLMs, attackers must manipulate the original image so that VLMs recognize only the specific target as changed while preserving the interpretation of the untargeted objects. In the example shown in \cref{fig:motivation_vs_baselines}, the image contains books, alphabet blocks, pencils, a drawing, and an apple. In this example, the target is \textbf{the apple} in the image, and the prompt is \textbf{a baseball}. This ensures that only the intended target is altered, while the overall context and responses to other parts of the input remain unchanged.

In previous studies, the authors of \cite{lu2023set} have changed a specific part of an output caption by introducing an adversarial perturbation. However, this method only alters a specific output token to another targeted token, limiting its generalization capabilities. Another approach involves updating the visual encoder~\cite{zhou2023advclip,zhao2023on}, which shares the same semantic space with a text encoder. This method perturbs images using text-guided features. In \cite{ye2024mutual}, a similar approach has been proposed, where the focus is on multi-modal architectures to enhance adversarial attacks.

Additionally, adversarial perturbations can be crafted by targeting the interactions between image and text modalities. Several studies have investigated adversarial attacks on contrastive language-image pre-training (CLIP)-like models, utilizing gradient-based optimization to create perturbations that disrupt multi-modal consistency~\cite{inkawhich2023adversarial, mao2022understanding}. These methods aim to misalign the shared semantic space between images and texts, causing the model to generate incorrect outputs.
However, despite numerous studies, none has addressed visual reasoning in adversarial attacks.




\paragraph*{Challenge 2---Effective Adversarial Example Algorithms}

Adversarial examples in VLMs pose a unique challenge, especially since models such as CLIP, which are multi-modal in nature, have shown inherent robustness against adversarial attacks due to their cross-modal learning ability~\cite{schlarmann2024robust}. CLIP models, by mapping images and texts into a shared semantic space, make it harder to generate adversarial perturbations that consistently fool both the vision and text encoders. This robustness stems from the model's ability to align multi-modal features, which adds an extra layer of complexity for adversarial attacks.

In previous studies~\cite{goodfellow2014explaining, dong2018boosting, lin2019nesterov, xie2019improving, wang2021enhancing}, several adversarial attack algorithms have been proposed; however, these algorithms focus on other issues of adversarial attacks---namely, transferability. To the best of the authors' knowledge, while multi-modal models like CLIP have demonstrated strong adversarial robustness, continued research is necessary to develop more effective adversarial example algorithms that can break this resilience. This includes refining techniques for creating targeted attacks that perturb the model's understanding of a specific visual object without affecting unrelated parts of the input.


\subsection{Summary of Our Contributions}

To address these challenges, we propose generating targeted adversarial examples through two steps: 1) applying a mask to the relevant part of the image corresponding to the target object, and 2) altering the masked region based on the desired prompt. This process creates a target image for adversarial attacks, where the generated examples visually resemble the original image but are recognized by the model as containing the intended synthesized object.

Our key contributions in this work are as follows:

\begin{itemize}
    \item \textbf{Replace-then-Perturb:} We propose a novel framework for generating targeted adversarial examples for VLMs, ensuring that only the specified target is altered while maintaining the integrity of the rest of the input.
    \item \textbf{Contrastive-Adv:} We introduce an adv aboveersarial gradient descent algorithm tailored for VLMs.
    \item \textbf{Benchmark Dataset Creation:} We develop a benchmark dataset for targeted adversarial examples in VLMs, providing a valuable resource for future research.
    \item \textbf{Comprehensive Evaluation:} We conduct extensive experiments using various adversarial example generation techniques to evaluate their effectiveness.
\end{itemize}

\subsection{Paper Organization}
The remainder of this paper is organized as follows.
Section \ref{sec:background} provides the background and preliminaries with a review of relevant literature and essential concepts. 
The proposed approach (\texttt{Replace-then-Perturb} and \texttt{Contrastive-Adv}) is described in Section III.
Section IV introduces the construction of our proposed dataset, TA-VLM. 
The experimental results in Section V present and analyzes the proposed method by comparing it with baselines. 
Finally, the conclusion summarizes the finding and suggests directions for future research.

\begin{figure*}
    \includegraphics[width=1\linewidth]{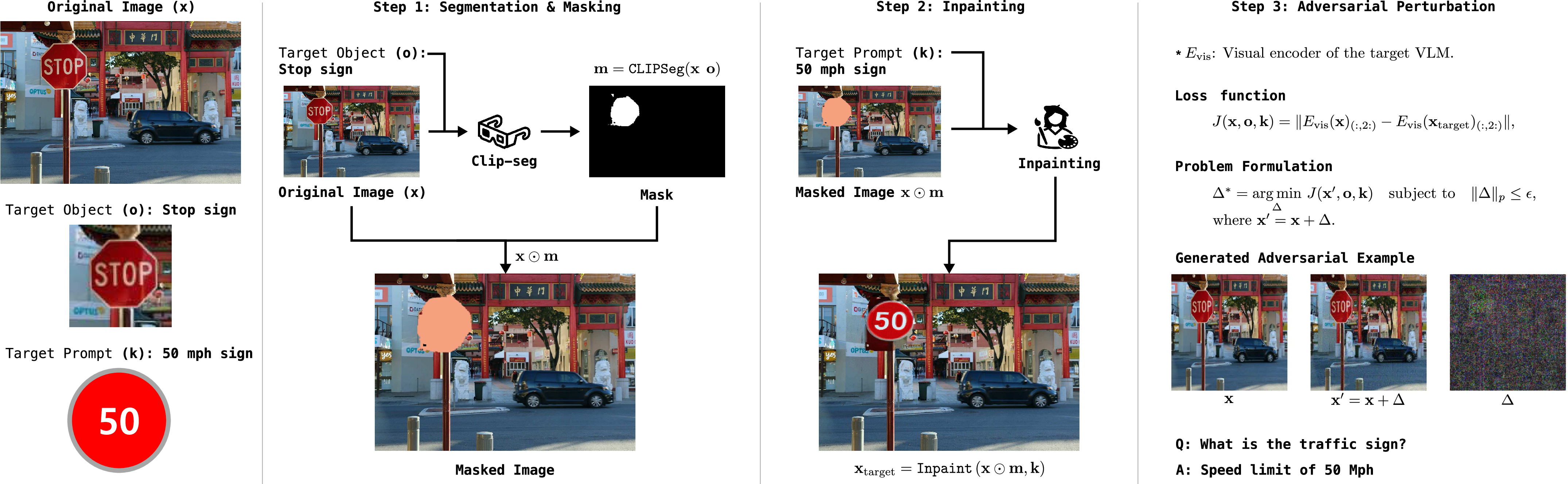}
    \caption{An illustration of the detailed procedure of \texttt{Replace-then-Perturb}, where the target object is the stop sign in the original image. In this adversarial attack, we aim to change the stop sign into a 50 mph speed limit sign. }
    \label{fig:RtP}
\end{figure*}

\section{Background and Preliminaries}
\label{sec:background}

In this section, we provide background information and preliminaries to contextualize our work. We first review existing adversarial attack algorithms. Next, we discuss the methodologies employed in prior studies. Experts and practitioners already familiar with VLM models and adversarial examples may proceed directly to the subsequent sections.



\paragraph*{Untargeted adversarial attacks}
As mentioned in the previous section, adversarial attacks are categorized into: 1) targeted adversarial attacks and 2) untargeted adversarial attacks.
In untargeted adversarial attacks, the aim is to cause a neural network model to misclassify the input image by adding a small perturbation to it.
For an original input image $\mathbf{x}$ with a ground-truth label $y$, the perturbed image, $\mathbf{x}' = \mathbf{x} + \Delta$, is crafted to maximize the loss function $J(\mathbf{x}', y, \theta)$, where the loss function could be any loss function such as mean square error (MSE) or cross entropy (CE).
Then, under a constraint on the maximum perturbation in the $p$-norm, the optimization problem for untargeted adversarial attacks is formulated as follows:
\begin{equation}\label{eq:untargeted}
\Delta^* = {\arg\max}_{\Delta} J(\mathbf{x}', y; \theta) \quad \text{subject to} \quad \Vert \Delta \Vert_\infty \leq \epsilon
\end{equation}
where $\epsilon$ is a constant defining the perturbation constraint.

\paragraph*{Targeted adversarial attacks}
Targeted adversarial attacks involve perturbing the input image to deceive neural network models. Unlike untargeted attacks, targeted adversarial attacks aim to force the model to predict a specific incorrect label, $y_{\text{target}}$. The optimization problem for targeted adversarial attacks is formulated as follows:
\begin{equation}\label{eq:targeted}
\Delta^* = {\arg\min}_{\Delta} J(\mathbf{x}', y_{\text{target}}, \theta) \quad \text{subject to} \quad \|\Delta\|_\infty \leq \epsilon
\end{equation}
where $\epsilon$ defines the maximum allowable perturbation.

\subsection{Adversarial Attack Algorithms}

To achieve the optimization goals of the problems in Equations \eqref{eq:untargeted} and \eqref{eq:targeted}, several methods have been proposed based on signed stochastic gradient descent.

\paragraph*{I-FGSM~\cite{goodfellow2014explaining}}
The Fast Gradient Sign Method (FGSM) is a popular and efficient approach for generating adversarial examples. To generate more effective adversarial examples, one can iteratively apply the FGSM algorithm, i.e., Iterative FGSM (I-FGSM). The update procedure of FGSM is represented by
\begin{equation}
\mathbf{x}_{t+1} = \textrm{Proj}_\epsilon \left(\mathbf{x}_t + \alpha \cdot \text{sign}\left(\nabla_{\mathbf{x}} J(\mathbf{x}, y; \theta)\right)\right)
\end{equation}
where $\alpha$ is a small constant that controls the step size, and $\text{sign}\left(\nabla_{\mathbf{x}} J(\mathbf{x}, y; \theta)\right)$ represents the sign of the gradient of the loss function with respect to the input. Moreover, the function $\text{Proj}_\epsilon(\cdot)$ denotes the projection of the input onto the space defined by $\Vert \Delta \Vert_\infty \leq \epsilon$.

\paragraph*{VMI-FGSM~\cite{wang2021enhancing}}

To further enhance the performance of the iterative FGSM, the Variance-tuned Momentum Iterative Fast Gradient Sign Method (VMI-FGSM) has been proposed. The main idea of this approach is to reduce the variance of the gradient, thereby stabilizing the update direction and helping to escape poor local optima. In each iteration, the variance reduction starts with:
\begin{equation}\label{eq:vmi-fgsm-1}
    v_t = \mathbb{E}_{\mathbf{r} \sim U(-(\beta \cdot \epsilon)^d, (\beta \cdot \epsilon)^d)}\left[\nabla_{\mathbf{x}} J(\mathbf{x}_t + \mathbf{r}, y; \theta)  - \nabla_{\mathbf{x}} J(\mathbf{x}_t, y; \theta)\right]
\end{equation}
where $\mathbf{r} \sim U(-(\beta \cdot \epsilon)^d, (\beta \cdot \epsilon)^d)$. We note that the vector $v_t$ is computed empirically, thereby requiring numerous iterations. Then, the variance-tuned momentum gradient can be obtained as:
\begin{equation}\label{eq:vmi-fgsm-2}
    \mathbf{g}_{t+1} = \mu \cdot \mathbf{g}_t + \frac{\nabla_{\mathbf{x}} J(\mathbf{x}_t, y; \theta) + v_t}{\Vert \nabla_{\mathbf{x}} J(\mathbf{x}_t, y; \theta) + v_t \Vert_1}
\end{equation}
where $\mu$ is a constant that controls the momentum weight. Finally, the perturbed image is updated by
\begin{equation}\label{eq:vmi-fgsm-3}
    \mathbf{x}_{t+1} = \mathrm{Proj}_{\epsilon}\left(\mathbf{x}_t + \alpha \cdot \mathrm{sign}(\mathbf{g}_{t+1})\right)
\end{equation}

Additionally, Momentum Iterative FGSM (MI-FGSM)~\cite{dong2018boosting}, Nesterov Iterative FGSM (NI-FGSM)~\cite{lin2019nesterov}, Projected Iterative FGSM (PI-FGSM)~\cite{xie2019improving}, and their extensions have been previously developed and are also considered in our experiments.

\subsection{Adversarial Attacks for VLM Models}

Here, we introduce existing approaches for generating adversarial examples against VLMs. Existing VLMs are categorized into: 1) late-fusion VLMs, \textit{i.e.}, models with separate visual and text encoders, and 2) early-fusion VLMs. In late-fusion VLMs, separate visual and text encoders process the visual and textual inputs into a shared latent space, respectively. Several VLMs leverage the late-fusion architecture, including Kosmos-2~\cite{peng2023kosmos}, the LLAVA families~\cite{liu2023improvedllava,liu2024llavanext,liu2024visual}, and Claude 3.5~\cite{claude2024}. On the other hand, few studies have proposed early-fusion VLMs, where visual and textual inputs are encoded into a unified encoder. However, due to the challenges of training foundation models without pre-trained encoders, only one such foundation model exists within this architecture~\cite{team2024chameleon}. As most VLMs utilize a late-fusion architecture, we focus on generating adversarial examples against late-fusion VLMs.

\paragraph*{Latent-space-based Adversarial Examples} 
In \cite{zhao2023on, zhou2023advclip}, the authors generated visual adversarial examples by leveraging the encoded latent space. Let $E_{\text{vis}}(\cdot)$ denote the visual encoder, \textit{e.g.}, CLIP, which encodes an input image into patch-wise embedding vectors. If the number of pixels is $(N_\text{h}, N_\text{w})$, then for an input image $\mathbf{x}$, the encoded embedding vector is represented by $E_{\text{vis}}(\mathbf{x}) \in \mathbb{R}^{N_f \times (N_\text{h} \cdot N_\text{w} + 1)}$, where $N_f$ denotes the feature dimension of the encoded features. This output consists of one CLS token and $(N_\text{h} \cdot N_\text{w})$ patch-wise feature vectors. The CLS token is used to project the visual input into the latent space shared with the text encoder. For instance, if we denote the first column of a matrix as $[\cdot]_{(:,1)}$, then the projected latent vector can be represented by $P_{\text{vis}}([E_{\text{vis}}(\mathbf{x})]_{(:,1)})$, where $P_{\text{vis}}$ represents the visual-to-latent space projection module. 

Let the target text of the adversarial example be denoted as $\mathbf{k}$ and its latent vector as $E_\text{text}(\mathbf{k})$. The latent-space-based adversarial example aims to minimize the following loss function:
\begin{equation}\label{eq:latent_based}
    J(\mathbf{x}, \mathbf{k}) = \Vert P_{\text{vis}}([E_{\text{vis}}(\mathbf{x})]_{(:,1)}) - E_\text{text}(\mathbf{k}) \Vert_2
\end{equation}

However, as depicted in \cref{fig:motivation_vs_baselines}, the latent-based approach does not effectively handle specific objects in the image, thereby allowing it to be detected by question-based adversarial example detection methods~\cite{ge2023mart}.

\section{Proposed Method}

In this section, we present our proposed method for generating targeted adversarial examples against visual-language multi-modal models.


\paragraph*{Our approach} 
As previously mentioned, targeted adversarial attacks aim to solve Problem \eqref{eq:targeted}. However, unlike previous studies~\cite{zhao2023on, zhou2023advclip}, we do not possess the target latent vector, target feature vector, or target class directly. Specifically, as illustrated in \eqref{eq:latent_based}, the target latent vector $E_{\text{text}}(\mathbf{k})$ does not effectively replace the target object with the target prompt. Instead of the approaches used in prior work, we aim to generate a new target feature vector $y_\text{target}$ (\texttt{Replace-then-Perturb}), where $y_\text{target}$ incorporates the latent vector with the target object replaced by the target prompt. Furthermore, to enhance the adversarial example generation process, we propose an efficient algorithm based on contrastive learning (\texttt{Contrastive-Adv}).

In the subsequent step, we denote the text representation of the target object as $\mathbf{o}$, and the target prompt as $\mathbf{k}$.  

\subsection{\texttt{Replace-then-Perturb}}

\paragraph*{Replace-the-Perturb} 
The \texttt{Replace-the-Perturb} method consists of three steps: (1) segmentation and masking, (2) inpainting, and (3) perturbation. In this subsection, we detail the first two steps (Replace steps), while the perturbation step will be introduced in the next section.

\paragraph*{1) Segmentation and Masking Step}

In the first step, we utilize a prompt-based segmentation model~\cite{luddecke2022image} to obtain the mask of the target object. As depicted in \cref{fig:RtP}, CLIP-based text and visual transformer models share the same latent space, thereby enabling zero-shot segmentation. In our approach, by forwarding a text prompt to this model, we obtain a binary mask $\mathbf{m} \in \{0,1\}^{w \times h}$, which has the same dimensions as the input image. 

Denoting the zero-shot segmentation model as $\texttt{CLIPSeg}$, the above procedure can be expressed as:
\begin{equation}\label{eq:CLIPSeg}
    \mathbf{m} = \mathtt{CLIPSeg}(\mathbf{x}, \mathbf{o}) \in \{0, 1\}^{w \times h},
\end{equation}
where the region of interest corresponding to the target object is marked as zero, and the background is marked as one.

For instance, in \cref{fig:RtP}, the target object is the `\textbf{Stop Sign}', and the target prompt is the `\textbf{50 Mph Sign}'. By performing zero-shot segmentation with the text `\textbf{Stop Sign}', we obtain the binary mask corresponding to the stop sign in the image.

\paragraph*{2) Inpainting Step}

In this stage, our focus is to replace the target object with the target prompt. For example, in \cref{fig:RtP}, the \texttt{Replace-the-Perturb} algorithm replaces the `\textbf{Stop Sign}' with the `\textbf{50 Mph Sign}'. Using the mask obtained in the previous stage, we mask out the target object as follows:
\begin{equation}\label{eq:masking}
    \hat{\mathbf{x}} = \mathbf{m} \odot \mathbf{x},
\end{equation}
where $\odot$ denotes the Hadamard product of two matrices or vectors with the same dimensions.

Next, by leveraging a prompt-based image inpainting method~\cite{kandinsky_22,Rombach_2022_CVPR,DeepFloydIF}, we fill the unmasked region of the image with the given target prompt. Denoting the prompt-based image inpainting model as $\mathtt{Inpaint}$, the revised image with the target prompt $\mathbf{k}$ is expressed as:
\begin{equation}\label{eq:feature_target}
    \mathbf{x}_{\text{target}} = \mathtt{Inpaint}\left(\mathbf{x} \odot \mathbf{m}, \mathbf{k}\right)
\end{equation}

In \cref{fig:RtP}, an example of the revised image via \texttt{Replace-the-Perturb} is depicted, where the region of the target object `\textbf{Stop Sign}' is masked and inpainted with the target prompt `\textbf{50 Mph Sign}'. 

For the adversarial perturbation, we utilize the feature vector output of the visual encoder as follows:
\begin{equation}\label{eq:feature_target_2}
    \mathbf{f}_{\text{target}} = E_\text{vis}(\mathbf{x}_{\text{target}})_{(:,2:)}
\end{equation}
where $\mathbf{X}_{(:,2:)}$ denotes the second to last columns of matrix $\mathbf{X}$. As mentioned earlier, the first column of the feature vector $E_\text{vis}(\mathbf{x})$ denotes the \texttt{CLS} token, which is used for obtaining the latent vector projected into the shared latent space. In our approach, instead of utilizing the projected latent vector, we use the patch-wise feature vectors of the input image. As depicted in \cref{fig:RtP}, the proposed method can generate a target feature vector supporting \textbf{visual reasoning}. The overall procedure of the \texttt{Replace-then-Perturb} method is summarized in Algorithm \ref{alg:RtP}.

\begin{algorithm}[!t]
\caption{\texttt{Replace-then-Perturb}$(\mathbf{x},\mathbf{o},\mathbf{k})$}
\label{alg:RtP}
\begin{algorithmic}[1] 
\STATE \textbf{Input}: Input image $\mathbf{x}$, target object $\mathbf{o}$, target prompt $\mathbf{k}$\\
\STATE \textbf{Step 1: Segmentation. }~Obtain mask for the target object $\mathbf{m}$ via Eq. \eqref{eq:CLIPSeg}. 
\STATE \textbf{Step 2: Replace. }~Obtain revised image $\mathbf{x}_\text{target}$ from Eq. \eqref{eq:feature_target}.
\IF {Perturbation algorithm is \texttt{Contrastive-Adv}}
    \STATE \textbf{Step 3: Perturb. }~Run \texttt{Contrastive-Adv}$(\mathbf{x},\mathbf{x}_\text{target})$ in Algorithm \ref{alg:contrastive-adv}.
\ELSE
    \STATE \textbf{Step 3: Target feature. }~Obtain the target feature $\mathbf{f}_\text{target}$ from Eq. \eqref{eq:feature_target_2}.
    \STATE \textbf{Step 4: Perturb. }~Solve Problem \eqref{eq:targeted_proposed} using adversarial example algorithms such as \texttt{FGMS} and \texttt{VMI-FGSM}.
\ENDIF
\STATE \textbf{Output}: Perturbation $\Delta$ and adversarial image $\mathbf{x}'$.
\end{algorithmic}
\end{algorithm}

Then, by leveraging we can configure the desired loss function for the adversarial example generation as 
\begin{equation}
    J(\mathbf{x}, \mathbf{o}, \mathbf{k}) = \Vert E_\text{vis}(\mathbf{x})_{(:,2:)} - E_\text{vis}(\mathbf{x}_{\text{target}})_{(:,2:)} \Vert,
\end{equation}
where $\mathbf{x}_\text{target}$ is obtained from \eqref{eq:feature_target}. 
Then, we aim to solve the following problem:
\begin{equation}\label{eq:targeted_proposed}
\Delta^* = \underset{\Delta}{\arg\min} ~ J(\mathbf{x}', \mathbf{o}, \mathbf{k}) \quad \text{subject to} \quad \|\Delta\|_p \leq \epsilon,
\end{equation}
where $\mathbf{x}'=\mathbf{x} + \Delta$. 

\subsection{\texttt{Contrastive-Adv}}

In this section, we propose an adversarial example generation algorithm inspired by contrastive learning. 
In the previous section, we propose a method to find the target feature vector $\mathbf{f}_\text{target}$. 
Then, the only remaining step is to get the perturbation matrix $\Delta^*$ by solving the problem \eqref{eq:targeted_proposed}. 
Existing adversarial example algorithms like VMI-FGSM~\cite{wang2021enhancing} can be used to make adversarial perturbations against VLMs. 
However, as discussed in~\cite{schlarmann2024robust}, VLMs have inherent robustness against adversarial examples due to their cross-modal learning ability.
In the numerical results, the benchmark verifies that the existing approaches cannot effectively generate adversarial examples against VLMs.

\begin{figure}
    \includegraphics[width=1\linewidth]{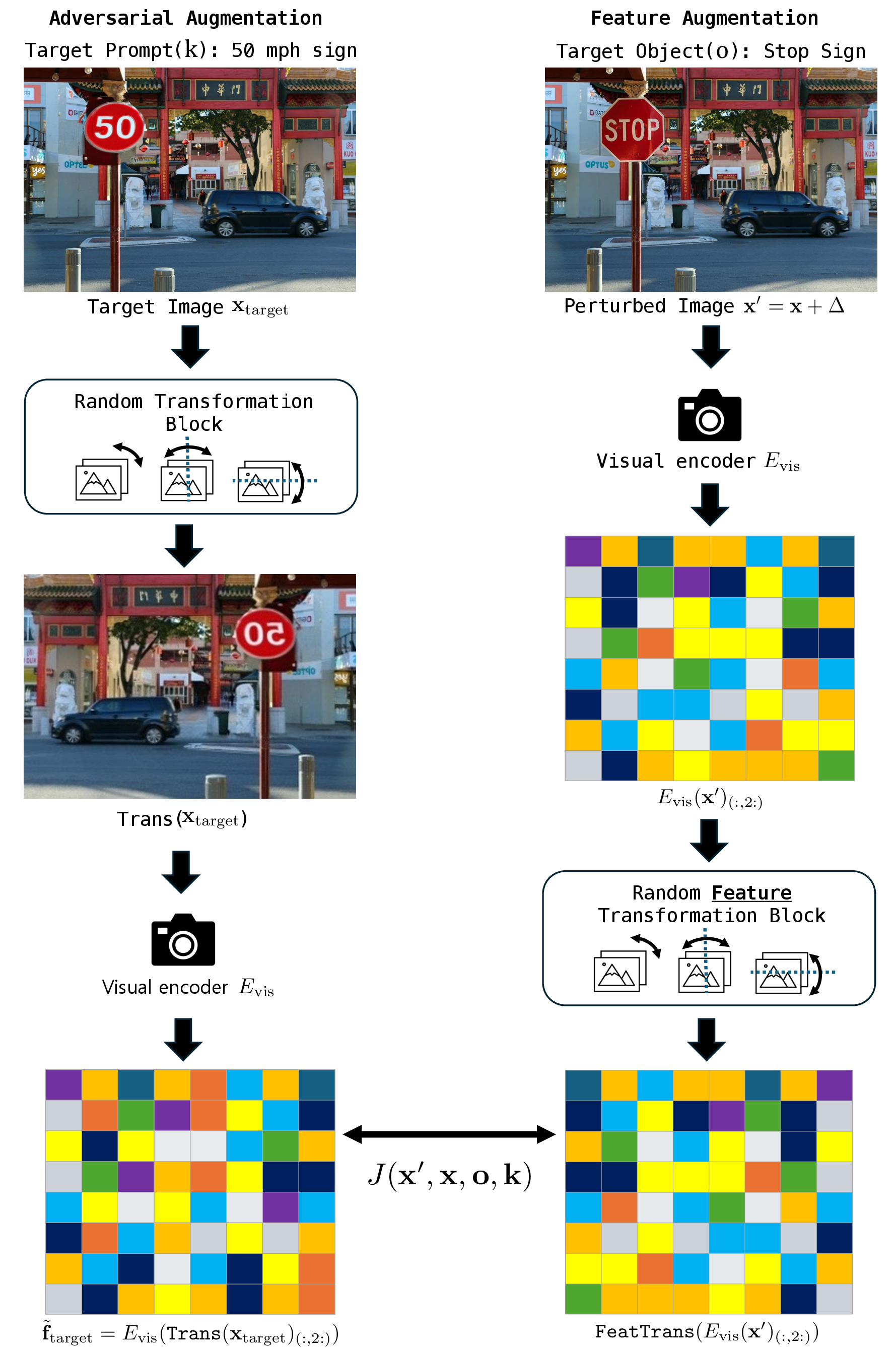}
    \caption{\texttt{Contrastive-Adv} algorithm}
    \label{fig:contrastive-adv}
\end{figure}

In \texttt{Contrastive-Adv} algorithm, we aim to escape from poor local optima and to stabilize the update direction with data augmentation and triplet loss function. 

\paragraph*{Adversarial augmentation}

In the first stage, we apply random image transformations such as random rotation, flip, and resize. 
Let us denote the random transformation block as $\mathtt{Trans}(\cdot)$. 
Then, the feature vector output of the visual encoder in \eqref{eq:feature_target_2} is newly defined by 
\begin{equation}\label{eq:f_tilde}
    \tilde{\mathbf{f}}_{\text{target}} = E_\text{vis}(\mathtt{Trans}(\mathbf{x}_{\text{target}})_{(:,2:)}).
\end{equation}
An example of the function $\mathtt{Trans}$ is depicted in \cref{fig:contrastive-adv}. 

\paragraph*{Feature augmentation}

In \eqref{eq:f_tilde}, we apply random image transformation. 
Because the outputs of the visual encoder $E_\text{vis}$ are patch-wise feature vectors, the feature vectors of the perturbed image $E_\text{vis}(\mathbf{x}')_{(:,2:)}$ are needed to be transformed via the same way.
To this end, we apply feature transformation as depicted in \cref{fig:contrastive-adv}. 
We transform the patch-level feature vectors with the same image transformation used in \eqref{eq:f_tilde}.
Denoting the feature transform function as $\mathtt{FeatTrans}$, the transformed patch-level feature vectors are denoted as 
\begin{equation}\label{eq:x_feature_filde}
    \mathtt{FeatTrans}(E_\text{vis}(\mathbf{x}')_{(:,2:)}).
\end{equation}

\begin{remark}[Why we do not apply augmentation block to the perturbation image?]
    The reason why we apply feature augmentation is that 
    But, it can have more general resistivity against adversarial purification methods~\cite{nie2022diffusion}
    However, our focus is not on developing a robust one; thus, we apply transformation blocks to the patch-level feature vectors of the perturbed image. 
\end{remark}

\paragraph*{Loss function}

We aim to minimize the gap between two vectors with the transformed feature vectors \eqref{eq:f_tilde} and \eqref{eq:x_feature_filde}. 
In addition to the $\ell$-2 loss function, apply the triplet loss function of the original image as follows:
\begin{equation}
\begin{split}
& J(\mathbf{x}', \mathbf{x}, \mathbf{o}, \mathbf{k}) = \\ 
& \Vert     \mathtt{FeatTrans}(E_\text{vis}(\mathbf{x}')_{(:,2:)}) -  E_\text{vis}(\mathtt{Trans}(\mathbf{x}_{\text{target}})_{(:,2:)}) \Vert \\
&- \mu \Vert     \mathtt{FeatTrans}(E_\text{vis}(\mathbf{x}')_{(:,2:)}) -  E_\text{vis}(\mathtt{Trans}(\mathbf{x})_{(:,2:)}) \Vert,
\end{split}
\end{equation}
where $\mu$ denotes the weight corresponding to the negative loss function. 

\begin{algorithm}[!t]
\caption{Iterative \texttt{Contrastive-Adv}$(\mathbf{x},\mathbf{x}_\text{target})$ with the standard FGSM}
\label{alg:contrastive-adv}
\begin{algorithmic}[1] 
\STATE \textbf{Input}: Original image $\mathbf{x}$, target image $\mathbf{x}_\text{target}$,  \\
\STATE  \textbf{Parameters}: Total iteration $T$, step size $\alpha$, and the maximum $\ell_\infty$ perturbation $\epsilon$.
\STATE \textbf{Initialize} $\Delta=\mathbf{0}$
\FOR {$t\in\{0,...,T-1\}$}
    \STATE Resample random transformation functions $\mathtt{Trans}$ and $\mathtt{FeatTrans}$. 
    \STATE $\mathbf{x}'\leftarrow\mathbf{x} + \Delta$
    \STATE Obtain the target feature vector outputs via \eqref{eq:f_tilde}. 
    \STATE Obtain the original feature vector output $E_\text{vis}(\mathtt{Trans}(\mathbf{x})_{(:,2:)})$.
    \STATE Compute the loss function $J(\mathbf{x}',\mathbf{x})$.
    \STATE $\mathbf{g}\leftarrow\nabla_{\mathbf{x}'} J(\mathbf{x}',\mathbf{x})$.
    \STATE $\bar{\mathbf{g}}\leftarrow \mathtt{sign}(\mathbf{g})$.
    \STATE $\Delta \leftarrow \Delta - \alpha \cdot \bar{\mathbf{g}}$
    \STATE $\Delta_i \leftarrow \mathtt{Clip}(\Delta_i, -\epsilon, \epsilon)$.\algorithmiccomment{Clipping}
\ENDFOR
\STATE \textbf{Output}: Perturbation $\Delta$.
\end{algorithmic}
\end{algorithm}

The step-by-step implementation of \texttt{Contrastive-Adv} is summarized in Algorithm \ref{alg:contrastive-adv}. 

\section{New Dataset and Quantitative Metrics}

\subsection{TA-VLM Dataset}

Current datasets, such as VQA, often face challenges in controlling unintended variations when altering the target object, resulting in inconsistent responses.

In this paper, we introduce the \texttt{TA-VLM} dataset, specifically designed to overcome limitations present in existing VQA datasets. 
The \texttt{TA-VLM} dataset addresses this issue by providing a controlled environment that ensures consistency in query responses. We employ this dataset to benchmark our proposed algorithm, demonstrating its effectiveness in managing these controlled scenarios.
The \texttt{TA-VLM} dataset comprises 100 natural images, each paired with approximately 10 queries, totaling 1,001 queries. 
Of these queries, 502 are classified as positive, while the remaining 499 are negative. Each image is associated with a target object and a corresponding target prompt. Positive queries are crafted to yield the same answer irrespective of the presence of the target object, whereas negative queries are designed so that the answer changes when the target object is modified based on the target prompt.

\subsection{Quantitative Metrics}

To evaluate performance on the \texttt{TA-VLM} dataset, we introduce a novel benchmarking metric. Unlike traditional VQA metrics, our metric does not rely on human annotations. Instead, we utilize responses from VLMs, enabling majority voting for more accurate measurements. We conducted experiments using five VLMs: LLAVA 1.5, LLAVA 1.6, Chameleon, Kosmos, and GPT-4o to provide answers for the images. Additionally, similarity models such as All-MINILM-v6~\cite{reimers-2019-sentence-bert}, BGE-M3~\cite{bge-m3}, and CLIP~\cite{hessel2021clipscore} are employed to evaluate the quality of the answers.

Given a target object query, the answers from the evaluation VLMs for images $\mathbf{x}$ and $\mathbf{x}_\text{target}$ are denoted as $\mathtt{ans}$ and $\mathtt{ans}_\text{target}$, respectively. The answer of the target VLM for the adversarial image $\mathbf{x}'$ is defined as $\mathtt{ans}_\text{adv}$. We compute the similarity score using semantic similarity models such as All-MINILM-v6~\cite{reimers-2019-sentence-bert}, BGE-M3~\cite{bge-m3}, and CLIP~\cite{hessel2021clipscore}. The similarity score determines whether the adversarial answer $\mathtt{ans}_\text{adv}$ is closer to the target answer $\mathtt{ans}_\text{target}$ compared to the original answer $\mathtt{ans}$. The detailed formula for the similarity score is as follows:
\begin{equation}
\text{score}(\mathtt{ans}, \mathtt{ans}_\text{target}, \mathtt{ans}_\text{adv}) = 
\begin{cases} 
1, & \text{if } \cos(\mathtt{ans}_\text{target}, \mathtt{ans}_\text{adv}) 
 \\ 
 & ~~~ \geq \cos(\mathtt{ans}, \mathtt{ans}_\text{adv}) \\
0, & \text{otherwise}.
\end{cases}    
\end{equation}

This metric assigns a binary score of 1 if the adversarial answer is semantically closer to the target answer than the original answer, and 0 otherwise. By aggregating these scores across all queries and models, we obtain an overall performance measure for our proposed adversarial example generation method.

\begin{figure*}
    \includegraphics[width=1\linewidth]{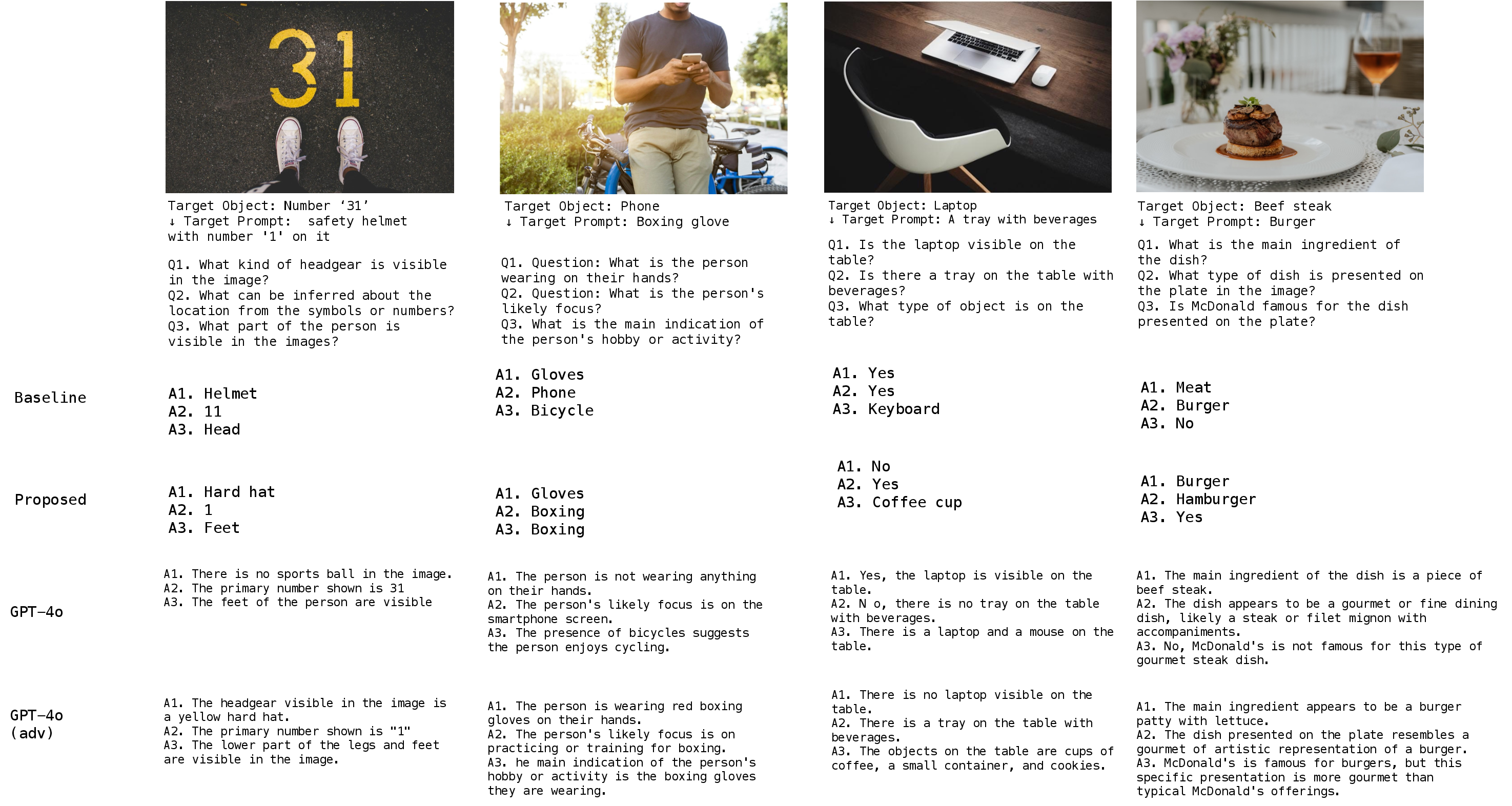}
    \caption{Graphical examples of the experimental results, where the target neural network model is LLAVA 1.5. 
    In this figure, we depict four examples, where the target object and the target prompt are indicated below the images.
    Based on the results, the latent-based adversarial examples (baseline) cannot consider the visual reasoning related to the adversarial changes. 
    In the first example, for the third question (\textbf{Q3}), the baseline answers that there is the head of a person in the image, as the target prompt `helmet' makes the perturbed image deceive the target model to recognize there is a person wearing a helmet. 
    In the second to fourth images, the adversarial examples generated by the baseline often lead VLMs to interpret the image in an unintended direction, whereas the proposed method accurately modifies the target of the adversarial perturbation. 
    }
    \label{fig:graphical_results}
\end{figure*}

\paragraph*{Majority voting}

We also employ a majority voting mechanism to combine predictions from different VLM models and enhance accuracy. Let $m_i$ represent the score of the $i$-th model, where $i = 1, 2, \dots, N$, and $N$ is the total number of models. The final prediction $\hat{m}$ is determined by the majority of the models' predictions, formulated as:
\begin{equation}\label{eq:majority_voting}
    \hat{m} = 
    \begin{cases} 
    1, & \text{if } \sum_{i=1}^{N} m_i \geq \frac{N}{2}, \\
    0, & \text{otherwise}.
    \end{cases}    
\end{equation}

In this context, $m_i \in \{0, 1\}$ indicates whether the $i$-th model's prediction is considered correct (1) or incorrect (0). The majority voting approach enhances the evaluation results by leveraging the collective decision-making capabilities of multiple models, thereby improving the overall accuracy and robustness of the assessment.

\section{Experiments}

In this section, we conduct a series of experiments to evaluate the adversarial examples against VLMs using the \texttt{TA-VLM} dataset.

\subsection{Experimental Details}

Our experimental setup utilizes a single NVIDIA RTX 3090 GPU. We configure the LLAVA series VLMs, specifically LLAVA 1.5 and LLAVA 1.6, as the target neural network models. The adversarial example algorithms are implemented for 200 steps, with the step size $\alpha$ fixed at $1/255$. In the \texttt{Contrastive-Adv} method, the weight $\mu$ is set to 0.3. The maximum perturbation is constrained using the $\ell_\infty$-norm, with values configured at 4/255, 8/255, 16/255, 32/255, and 64/255.

\subsection{Baselines and Metrics}

\paragraph*{Baselines}
In our evaluation, we compare our proposed methods against several established adversarial attack techniques. 
\begin{itemize}
    \item For the comparison with \texttt{Replace-then-Perturb}, we implement the latent-space-based adversarial example~\cite{zhou2023advclip}. 
    \item Also, to show superiority of \texttt{Contrastive-Adv}, we conduct the following baselines: FGSM~\cite{goodfellow2014explaining}, MI-FGSM~\cite{dong2018boosting}, NI-FGSM~\cite{lin2019nesterov}, PI-FGSM~\cite{xie2019improving}, VMI-FGSM~\cite{wang2021enhancing}. 
\end{itemize}

\paragraph*{Metrics}

Our benchmark employs both algorithm-based metrics and VLM-based metrics to evaluate the performance of adversarial examples on the \texttt{TA-VLM} dataset.
We note that images in the \texttt{TA-VLM} dataset are fully utilized for evaluation.
\begin{itemize}
    \item \textbf{Algorithm-based Metrics:} We utilize the Bilingual Evaluation Understudy (BLEU) score~\cite{papineni2002bleu} and the Generalized Language Evaluation Understanding (GLEU) metric~\cite{napoles2015ground}. These word-counting-based metrics assess the overlap of words between two given sentences, providing a quantitative measure of similarity.
    \item \textbf{VLM/LLM-based Metrics:} To capture the semantic similarity and contextual relevance of the answers, we employ sentence similarity models, including All-MINILM-v6~\cite{reimers-2019-sentence-bert}, BGE-M3~\cite{bge-m3}, and CLIP~\cite{hessel2021clipscore}. These models evaluate the semantic alignment between the adversarial answers and the target answers, offering a more nuanced assessment of answer quality beyond mere word overlap.    
\end{itemize}

\begin{table*}[!ht]
    \caption{Experimental results of the proposed method and baselines in algorithm-based metrics, where $\epsilon=16.0/255$ and $T=200$. }
    \centering
    \adjustbox{width=1\linewidth}{
    \begin{tabular}{c|c|c|ccccccc}
    \toprule
    \multirow{2}{*}{Evaluation VLM} & \multirow{2}{*}{Metrics} & \multirow{2}{*}{Latent-based~\cite{zhou2023advclip}}   &  \multicolumn{7}{c}{Replace-then-Perturb \textbf{(Ours)}} \\
    & & & I-FGSM & PI-FGSM & PI-FGSMPP & NI-FGSM & SINI-FGSM & VMI-FGSM & Contrastive-Adv \textbf{(Ours)} \\
    \midrule
    \multirow{2}{*}{Chameleon} & BLUE & 0.492  & 0.514  & 0.493  & 0.514  & 0.487  & 0.488  & \textbf{0.513}  & \textbf{0.513}  \\
    & GLUE & 0.499  & \textbf{0.517}  & 0.489  & 0.520  & 0.496  & 0.489  & 0.516  & 0.509  \\
    \multirow{2}{*}{Kosmos} & BLUE & 0.505  & 0.491  & 0.480  & 0.490  & 0.473  & 0.476  & 0.501  & \textbf{0.513}  \\
    & GLUE & 0.504  & 0.485  & 0.484  & 0.489  & 0.473  & 0.480  & 0.500  & \textbf{0.507}  \\
    \multirow{2}{*}{LLAVA 1.5} & BLUE & 0.477  & 0.511  & 0.382  & 0.542  & 0.310  & 0.312  & 0.557  & \textbf{0.633}  \\
    & GLUE & 0.470  & 0.506  & 0.377  & 0.538  & 0.304  & 0.307  & 0.557  & \textbf{0.631}  \\
    \multirow{2}{*}{LLAVA 1.6} & BLUE & 0.476  & 0.483  & 0.425  & 0.496  & 0.407  & 0.410  & 0.513  & \textbf{0.523}  \\
    & GLUE & 0.475  & 0.484  & 0.423  & 0.504  & 0.408  & 0.403  & 0.514  & \textbf{0.529}  \\
    \multirow{2}{*}{GPT 4o} & BLUE & 0.464  & 0.465  & 0.428  & 0.483  & 0.395  & 0.403  & 0.501  & \textbf{0.518}  \\
    & GLUE & 0.463  & 0.465  & 0.421  & 0.478  & 0.389  & 0.398  & 0.495  & \textbf{0.526}  \\ \midrule
    \multirow{2}{*}{majority vote} & BLUE & 0.607  & 0.625  & 0.491  & 0.657  & 0.391  & 0.411  & 0.671  & \textbf{0.768}  \\
     & GLUE & 0.609  & 0.625  & 0.487  & 0.655  & 0.393  & 0.405  & 0.671  & \textbf{0.768}  \\ \midrule 
    \multirow{2}{*}{Avg} & BLUE & 0.504  & 0.515  & 0.450  & 0.530  & 0.410  & 0.416  & 0.543  & \textbf{0.578}  \\
     & GLUE & 0.503  & 0.514  & 0.447  & 0.531  & 0.410  & 0.413  & 0.542  & \textbf{0.578}  \\
    \bottomrule
    \multicolumn{10}{l}{*The best scheme is highlighted by \textbf{bold}.}
    \end{tabular}
    }
    \label{tab:algorithm-based-metrics}
\end{table*}

\subsection{Graphical Results}

Figure \ref{fig:graphical_results} presents four illustrative examples demonstrating the effectiveness of our adversarial example generation method compared to the baseline latent-space-based approach.

In the first example, the adversarial perturbation aims to change the number `31' to a safety helmet bearing the number `1'. To achieve this, the baseline method adds a random perturbation to the original image, causing the visual encoder of the VLMs to align with the target prompt (`helmet'). Consequently, both the baseline and our proposed method produce answers directly related to the target prompt (\textbf{Q1}).
However, when the questions require visual reasoning, the baseline method often fails to provide correct answers (\textbf{Q3}). Specifically, Question \textbf{Q3} inquires about the visible part of the person in the image. Since the baseline method aligns the latent output with the target prompt, it erroneously infers that there is a person's head beneath the helmet. This unintended effect results in unnatural outputs. In contrast, our proposed method, which incorporates the background around the target object, is able to provide accurate and natural answers to such questions.

Similarly, in the other examples, the baseline method fails to effectively replace the target objects with the target prompts. For instance, when replacing the phone with boxing gloves, the baseline incorrectly indicates that the man still focuses on the phone, whereas our proposed method accurately reports his concentration on boxing gloves (\textbf{Q1}). In the third example, the baseline is unable to remove the laptop, leading the VLM to mention the keyboard, while our method successfully directs the VLM towards the intended replacement (\textbf{Q3}). The fourth example exhibits similar results, where the baseline correctly answers \textbf{Q2} based on the original image but fails to update responses to other questions, unlike our approach.

\paragraph*{Summary}

The graphical results demonstrate that our proposed method effectively challenges VLMs by accounting for complex relationships between objects and prompts. Unlike the baseline approach, our method offers a more robust and nuanced framework for adversarial evaluation, ensuring VLMs are tested against a broader range of challenging and contextually relevant perturbations.

\begin{table}
    \caption{Experimental results of the proposed method and baseline methods in algorithmic metrics (BLEU and GLEU) for various values of $\epsilon$. For the brevity of the presentation, we present the majority vote and average score of five different evaluation VLM models (Chameleon, Kosmos, LLAVA 1.5, LLAVA 1.6, GPT-4o). `RtP' is an abbreviation of Replace-then-Perturb. }
    \centering
    \adjustbox{width=1\linewidth}{
    \begin{tabular}{c|ccccc}
    \toprule
     & \multicolumn{5}{c}{Avg. BLEU Score (Majority vote)}\\
     & $\epsilon=4/255$ & $\epsilon=8/255$ & $\epsilon=16/255$ & $\epsilon=32/255$ & $\epsilon=64/255$ \\ 
    \midrule
    \makecell{I-FGSM \\ (Latent-based)}  & 0.429 (0.435) & 0.482 (0.553) & 0.504 (0.607) & 0.513 (0.633) & 0.508 (0.623) \\ 
    \midrule
    I-FGSM (RtP) & 0.410 (0.383) & 0.461 (0.505) & 0.515 (0.625) & 0.540 (0.669) & 0.545 (0.683) \\
    PI-FGSM (RtP)  & 0.401 (0.365) & 0.418 (0.415) & 0.450 (0.491) & 0.496 (0.593) & 0.529 (0.663) \\
    PI-FGSMPP (RtP)  & 0.403 (0.373) & 0.474 (0.535) & 0.530 (0.657) & 0.565 (0.725) & 0.578 (0.756) \\
    NI-FGSM (RtP)  & 0.384 (0.321) & 0.389 (0.335) & 0.410 (0.391) & 0.480 (0.539) & 0.531 (0.655) \\
    SINI-FGSM (RtP)  & 0.381 (0.313) & 0.392 (0.341) & 0.416 (0.411) & 0.501 (0.589) & 0.554 (0.703) \\
    VMI-FGSM (RtP)  & 0.418 (0.409) & 0.488 (0.563) & 0.543 (0.671) & 0.565 (0.729) & 0.562 (0.719) \\
    \textbf{Ours}  & \textbf{0.440 (0.457)} & \textbf{0.518 (0.649)} & \textbf{0.578 (0.768)} & \textbf{0.587 (0.772)} & \textbf{0.592 (0.794)} \\
    \toprule
    \toprule
     & \multicolumn{5}{c}{Avg. GLEU Score (Majority vote)}\\
     & $\epsilon=4/255$ & $\epsilon=8/255$ & $\epsilon=16/255$ & $\epsilon=32/255$ & $\epsilon=64/255$ \\ 
    \midrule
    \makecell{I-FGSM \\ (Latent-based)}  & 0.430 (0.437) & 0.480 (0.553) & 0.503 (0.609) & 0.514 (0.637) & 0.509 (0.627) \\
    \midrule
    I-FGSM (RtP)  & 0.406 (0.379) & 0.459 (0.503) & 0.514 (0.625) & 0.544 (0.679) & 0.544 (0.681) \\
    PI-FGSM (RtP)  & 0.400 (0.369) & 0.420 (0.423) & 0.447 (0.487) & 0.495 (0.593) & 0.533 (0.667) \\
    PI-FGSMPP (RtP)  & 0.401 (0.373) & 0.475 (0.541) & 0.531 (0.655) & 0.567 (0.731) & 0.583 (0.762) \\
    NI-FGSM (RtP)  & 0.381 (0.323) & 0.386 (0.333) & 0.410 (0.393) & 0.476 (0.535) & 0.531 (0.659) \\
    SINI-FGSM (RtP)  & 0.379 (0.315) & 0.389 (0.337) & 0.413 (0.405) & 0.497 (0.583) & 0.554 (0.703) \\
    VMI-FGSM (RtP)  & 0.418 (0.413) & 0.488 (0.565) & 0.542 (0.671) & 0.568 (0.733) & 0.563 (0.725) \\
    \textbf{Ours}  & \textbf{0.440 (0.465)} & \textbf{0.517 (0.645)} & \textbf{0.578 (0.768)} & \textbf{0.587 (0.774)} & \textbf{0.592 (0.796)} \\
    \bottomrule
    \multicolumn{6}{l}{*The best scheme is highlighted by \textbf{bold}.}
    \end{tabular}
    }
    \label{tab:various_epsilon}
\end{table}

\begin{table*}[!ht]
    \caption{Experimental results of the proposed method and baselines in VLM-based metrics, where $\epsilon=16.0/255$ and $T=200$. }
    \centering
    \adjustbox{width=1\linewidth}{
    \begin{tabular}{c|c|c|ccccccc}
    \toprule
    \multirow{2}{*}{Evaluation VLM} & \multirow{2}{*}{Metrics} & \multirow{2}{*}{Latent-based~\cite{zhou2023advclip}}   &  \multicolumn{7}{c}{Replace-then-Perturb \textbf{(Ours)}} \\
    & & & I-FGSM & PI-FGSM & PI-FGSMPP & NI-FGSM & SINI-FGSM & VMI-FGSM & Contrastive-Adv \textbf{(Ours)} \\
    \midrule
    \multirow{6}{*}{Chameleon} & ALL miniLM & 0.457  & 0.481  & 0.427  & 0.503  & 0.391  & 0.393  & 0.507  & \textbf{0.545}  \\
    & BGE-M3-COLBERT & 0.421  & 0.457  & 0.409  & 0.459  & 0.365  & 0.389  & 0.483  & \textbf{0.513}  \\
    & BGE-M3-DENSE & 0.404  & 0.451  & 0.400  & 0.456  & 0.354  & 0.366  & 0.474  & \textbf{0.508}  \\
    & BERT & 0.462  & 0.480  & 0.452  & 0.494  & 0.391  & 0.387  & 0.488  & \textbf{0.531}  \\
    & CLIP (image-txt) & 0.451  & 0.467  & 0.431  & 0.467  & 0.409  & 0.409  & 0.483  & \textbf{0.503}  \\
    & CLIP & 0.443  & 0.463  & 0.423  & 0.461  & 0.389  & 0.395  & 0.481  & \textbf{0.503}  \\
    \multirow{6}{*}{Kosmos} & ALL miniLM & 0.463  & 0.461  & 0.409  & 0.467  & 0.363  & 0.363  & 0.487  & \textbf{0.549}  \\
    & BGE-M3-COLBERT & 0.461  & 0.449  & 0.397  & 0.457  & 0.355  & 0.355  & 0.485  & \textbf{0.549}  \\
    & BGE-M3-DENSE & 0.451  & 0.467  & 0.397  & 0.471  & 0.353  & 0.355  & 0.493  & \textbf{0.553}  \\
    & BERT & 0.456  & 0.504  & 0.423  & 0.500  & 0.380  & 0.383  & 0.534  & \textbf{0.571}  \\
    & CLIP (image-txt) & 0.459  & 0.467  & 0.427  & 0.465  & 0.409  & 0.403  & 0.487  & \textbf{0.517}  \\
    & CLIP & 0.459  & 0.469  & 0.431  & 0.469  & 0.397  & 0.397  & 0.491  & \textbf{0.537}  \\
    \multirow{6}{*}{LLAVA 1.5} & ALL miniLM & 0.458  & 0.518  & 0.380  & 0.552  & 0.302  & 0.296  & 0.572  & \textbf{0.670}  \\
    & BGE-M3-COLBERT & 0.448  & 0.504  & 0.372  & 0.552  & 0.302  & 0.304  & 0.566  & \textbf{0.644}  \\
    & BGE-M3-DENSE & 0.454  & 0.509  & 0.372  & 0.549  & 0.304  & 0.294  & 0.557  & \textbf{0.647}  \\
    & BERT & 0.461  & 0.506  & 0.365  & 0.515  & 0.312  & 0.317  & 0.640  & \textbf{0.618}  \\
    & CLIP (image-txt) & 0.454  & 0.502  & 0.376  & 0.530  & 0.304  & 0.310  & 0.568  & \textbf{0.612}  \\
    & CLIP & 0.462  & 0.510  & 0.380  & 0.542  & 0.316  & 0.310  & 0.570  & \textbf{0.634}  \\
    \multirow{6}{*}{LLAVA 1.6} & ALL miniLM & 0.457  & 0.499  & 0.411  & 0.495  & 0.325  & 0.341  & 0.555  & \textbf{0.595} \\
    & BGE-M3-COLBERT & 0.447  & 0.507  & 0.393  & 0.515  & 0.319  & 0.311  & 0.541  & \textbf{0.595}  \\
    & BGE-M3-DENSE & 0.427  & 0.485  & 0.385  & 0.503  & 0.305  & 0.305  & 0.521  & \textbf{0.583}  \\
    & BERT & 0.434  & 0.507  & 0.372  & 0.507  & 0.310  & 0.307  & 0.557  & \textbf{0.605}  \\
    & CLIP (image-txt) & 0.465  & 0.485  & 0.433  & 0.485  & 0.403  & 0.401  & 0.507  & \textbf{0.541}  \\
    & CLIP & 0.469  & 0.511  & 0.435  & 0.501  & 0.377  & 0.381  & 0.535  & \textbf{0.571}  \\
    \multirow{6}{*}{GPT-4o} & ALL miniLM & 0.398  & 0.418  & 0.346  & 0.444  & 0.274  & 0.274  & 0.470  & \textbf{0.518}  \\
    & BGE-M3-COLBERT & 0.406  & 0.430  & 0.332  & 0.446  & 0.276  & 0.270  & 0.494  & \textbf{0.552}  \\
    & BGE-M3-DENSE & 0.392  & 0.421  & 0.342  & 0.434  & 0.273  & 0.265  & 0.480  & \textbf{0.548}  \\
    & BERT & 0.395  & 0.420  & 0.341  & 0.444  & 0.280  & 0.268  & 0.480  & \textbf{0.546}  \\
    & CLIP (image-txt) & 0.494  & 0.516  & 0.460  & 0.508  & 0.436  & 0.426  & 0.548  & \textbf{0.582}  \\
    & CLIP & 0.506  & 0.518  & 0.446  & 0.528  & 0.402  & 0.390  & 0.552  & \textbf{0.608}  \\
    \midrule
    \multirow{6}{*}{majority vote} & ALL miniLM & 0.473  & 0.509  & 0.409  & 0.525  & 0.323  & 0.323 & 0.561  & \textbf{0.631}  \\
    & BGE-M3-COLBERT & 0.467  & 0.509  & 0.389  & 0.531  & 0.311  & 0.321  & 0.569  &\textbf{0.639}  \\
    & BGE-M3-DENSE & 0.431  & 0.489  & 0.357  & 0.505  & 0.283  & 0.289  & 0.535  & \textbf{0.635}  \\
    & BERT & 0.479  & 0.515  & 0.375  & 0.497  & 0.303  & 0.297  & 0.573  & \textbf{0.625}  \\
    & CLIP (image-txt) & 0.507  & 0.535  & 0.455  & 0.543  & 0.405  & 0.407  & 0.589  & \textbf{0.637}  \\
    & CLIP & 0.519  & 0.547  & 0.459  & 0.555  & 0.389  & 0.389  & 0.589  & \textbf{0.651}  \\\midrule
    \multirow{6}{*}{avg} & ALL miniLM & 0.451  & 0.481  & 0.397  & 0.498  & 0.329  & 0.331  & 0.525  & \textbf{0.585}  \\
    & BGE-M3-COLBERT & 0.442  & 0.476  & 0.382  & 0.493  & 0.321  & 0.325  & 0.523  & \textbf{0.582}  \\
    & BGE-M3-DENSE & 0.426  & 0.470  & 0.375  & 0.486  & 0.312  & 0.312  & 0.510  & \textbf{0.579}  \\
    & BERT & 0.448  & 0.489  & 0.388  & 0.493  & 0.329  & 0.326  & 0.545  & \textbf{0.583}  \\
    & CLIP (image-txt) & 0.472  & 0.495  & 0.430  & 0.500  & 0.394  & 0.392  & 0.530  & \textbf{0.565}  \\
    & CLIP & 0.476  & 0.503  & 0.429  & 0.509  & 0.378  & 0.377  & 0.536  & \textbf{0.584}  \\
    \bottomrule
    \multicolumn{10}{l}{*The best scheme is highlighted by \textbf{bold}.}
    \end{tabular}
    }
    \label{tab:VLM-based-metrics}
\end{table*}

\subsection{Quantitative Results---Algorithm-Based Metrics}\label{subsec:quantitative_results_algorithm}

We demonstrate the superiority of our proposed methods using algorithm-based metrics. Table \ref{tab:algorithm-based-metrics} presents a detailed comparison of various adversarial attack methods applied to the target Visual-Language Model (VLM), LLAVA 1.5. To generate reference answers, we utilize responses from Chameleon, Kosmos, LLAVA 1.5, LLAVA 1.6, and GPT-4o. The table contrasts the baseline latent-space-based method with our \texttt{Replace-then-Perturb} method. Furthermore, we compare our \texttt{Contrastive-Adv} method against several established adversarial attack techniques, including FGSM~\cite{goodfellow2014explaining}, MI-FGSM~\cite{dong2018boosting}, PI-FGSM~\cite{xie2019improving}, and VMI-FGSM~\cite{wang2021enhancing}. We employ two key metrics, BLEU and GLEU, to assess model performance.

\paragraph*{Contrastive-Adv vs. Adversarial Attack Methods}
Based on the BLEU and GLEU scores for each VLM model, the baseline method performs adequately but is outperformed by our proposed methods across most models. Notably, for the Chameleon and LLAVA models, the proposed \texttt{Contrastive-Adv} method achieves higher BLEU and GLEU scores compared to existing adversarial attack methods. These results demonstrate that our method is more effective in generating adversarial examples for VLMs by leveraging the contrastive training procedures inherent in VLM training.

\paragraph*{Replace-then-Perturb vs. Latent-based}
In our ablation study, we compare the latent-based method and I-FGSM with the \texttt{Replace-then-Perturb} method, as shown in Table \ref{tab:algorithm-based-metrics}.\footnote{Note that the latent-based method also utilizes the I-FGSM algorithm for updating perturbation noise.} The \texttt{Replace-then-Perturb} method slightly outperforms the latent-based method in terms of BLEU and GLEU scores. This improvement is attributed to the fact that the latent-based method operates in a projected latent vector space that is significantly lower-dimensional compared to the patch-wise feature vectors, as illustrated in Equations \eqref{eq:latent_based} and \eqref{eq:feature_target_2}. Despite this marginal enhancement, the \texttt{Replace-then-Perturb} method can further improve its performance through our other proposed method, \texttt{Contrastive-Adv}, which is not applicable to latent-based adversarial attacks.

\paragraph*{Various values of $\epsilon$}
Table \ref{tab:various_epsilon} presents experimental results for different values of $\epsilon$, representing the maximum allowable perturbation per pixel. For example, with $\epsilon=16/255$, adversarial attackers can alter each pixel's value by up to 16/255. Generally, increasing $\epsilon$ enhances the performance of adversarial attacks. To maintain brevity, the table displays only the average and majority vote scores for the BLEU and GLEU metrics. Notably, our proposed method significantly outperforms all baseline methods. While the latent-based I-FGSM method benefits from the low-dimensional latent space and outperforms the Replace-then-Perturb-based I-FGSM at smaller $\epsilon$ values, our method consistently surpasses the latent-based approach across all $\epsilon$ levels.

\subsection{Quantitative Results---LLM/VLM-based metrics}

In this section, we assess the superiority of our proposed method using VLM-based metrics. Table \ref{tab:VLM-based-metrics} presents a detailed comparison between existing adversarial attack methods and our approach. We utilize five different VLMs--Chameleon, Kosmos, LLAVA 1.5, LLAVA 1.6, and GPT-4o--to generate reference answers and report both majority vote and average scores across these models. For evaluation, we leverage semantic similarity scores from models such as All-MiniLM, BGE-M3, BERT, and CLIP.

\paragraph*{Replace-then-Perturb vs. Latent-based}
For the ablation study, we compare the I-FGSM method using the latent-based approach against our \texttt{Replace-then-Perturb} method, as shown in the left two columns of Table \ref{tab:algorithm-based-metrics}. The proposed \texttt{Replace-then-Perturb} method outperforms the latent-based method, including methods based on CLIP, across all evaluation metrics.

\paragraph*{Contrastive-Adv vs. Other Adversarial Attacks}
The VLM-based scores in Table \ref{tab:VLM-based-metrics} reveal that the \texttt{Contrastive-Adv} method significantly outperforms baseline adversarial attacks using \texttt{Replace-then-Perturb}. For example, in the BGE-M3 dense score with majority voting, our method achieves a 10\% point higher score compared to the best baseline. Across all other metrics, \texttt{Contrastive-Adv} consistently surpasses all baseline methods. These results demonstrate that our approach effectively perturbs the original image, enabling the target VLM model to recognize the target object as the specified prompt.

\begin{table}
    \caption{
    Experimental results in VLM-based metrics for various values of $\epsilon$. For the brevity of the presentation, we present the majority vote and average scores for two representative metrics, BERT, CLIP, and BGE-M3. `RtP' is an abbreviation of Replace-then-Perturb.
    }
    \centering
    \adjustbox{width=1\linewidth}{
    \begin{tabular}{c|ccccc}
    \toprule
     & \multicolumn{5}{c}{Avg. BERT Score (Majority vote)}\\
     & $\epsilon=4/255$ & $\epsilon=8/255$ & $\epsilon=16/255$ & $\epsilon=32/255$ & $\epsilon=64/255$ \\ 
    \midrule
    \makecell{I-FGSM \\ (Latent-based)}    & 0.361 (0.345) & 0.400 (0.391) & 0.448 (0.479) & 0.455 (0.459) & 0.460 (0.483) \\
    \midrule
    I-FGSM (RtP)  & 0.340 (0.311) & 0.393 (0.383) & 0.489 (0.515) & 0.519 (0.543) & 0.529 (0.571) \\
    PI-FGSM (RtP)  & 0.320 (0.285) & 0.334 (0.289) & 0.388 (0.375) & 0.446 (0.451) & 0.475 (0.501) \\
    PI-FGSMPP (RtP) & 0.328 (0.289) & 0.422 (0.409) & 0.493 (0.497) & 0.553 (0.589) & 0.570 (0.617) \\
    NI-FGSM (RtP)   & 0.280 (0.230) & 0.308 (0.267) & 0.329 (0.303) & 0.422 (0.429) & 0.496 (0.515) \\
    SINI-FGSM (RtP) & 0.303 (0.269) & 0.290 (0.246) & 0.326 (0.297) & 0.452 (0.449) & 0.544 (0.575) \\
    VMI-FGSM (RtP)  & 0.339 (0.307) & 0.438 (0.437) & 0.545 (0.573) & 0.559 (0.599) & 0.554 (0.591) \\
    \textbf{Ours}   & \textbf{0.362 (0.347)} & \textbf{0.496 (0.539)} & \textbf{0.583 (0.625)} & \textbf{0.595 (0.649)} & \textbf{0.620 (0.677)} \\
    \toprule
    \toprule
     & \multicolumn{5}{c}{Avg. BGE-M3 Score (Majority vote)}\\
     & $\epsilon=4/255$ & $\epsilon=8/255$ & $\epsilon=16/255$ & $\epsilon=32/255$ & $\epsilon=64/255$ \\ 
    \midrule
    \midrule
    \makecell{I-FGSM \\ (Latent-based)}   & 0.329 (0.307) & 0.402 (0.397) & 0.426 (0.431) & 0.464 (0.487) & 0.445 (0.453) \\
    \midrule
    I-FGSM (RtP)  & 0.311 (0.279) & 0.389 (0.385) & 0.470 (0.489) & 0.505 (0.533) & 0.507 (0.543) \\
    PI-FGSM (RtP)   & 0.294 (0.257) & 0.327 (0.291) & 0.375 (0.357) & 0.429 (0.445) & 0.491 (0.517) \\
    PI-FGSMPP (RtP)  & 0.302 (0.267) & 0.406 (0.407) & 0.486 (0.505) & 0.546 (0.591) & 0.576 (0.629) \\
    NI-FGSM (RtP)   & 0.276 (0.232) & 0.280 (0.236) & 0.312 (0.283) & 0.416 (0.425) & 0.487 (0.513) \\
    SINI-FGSM (RtP)  & 0.275 (0.232) & 0.280 (0.230) & 0.312 (0.289) & 0.440 (0.449) & 0.538 (0.585) \\
    VMI-FGSM (RtP)   & 0.325 (0.295) & 0.427 (0.429) & 0.510 (0.535) & 0.551 (0.595) & 0.544 (0.581) \\
    \textbf{Ours}  & \textbf{0.352 (0.337)} & \textbf{0.468 (0.483)} &\textbf{ 0.579 (0.635)} & \textbf{0.591 (0.651)} & \textbf{0.604 (0.669)} \\
    \toprule
    \toprule
     & \multicolumn{5}{c}{Avg. CLIP Score (Majority vote)}\\
     & $\epsilon=4/255$ & $\epsilon=8/255$ & $\epsilon=16/255$ & $\epsilon=32/255$ & $\epsilon=64/255$ \\ 
    \midrule
    \midrule
    \makecell{I-FGSM \\ (Latent-based)}   & 0.385 (0.397) & 0.449 (0.485) & 0.476 (0.519) & 0.497 (0.543) & 0.490 (0.535) \\
    \midrule
    I-FGSM (RtP)  & 0.373 (0.381) & 0.437 (0.469) & 0.503 (0.547) & 0.529 (0.587) & 0.529 (0.589) \\
    PI-FGSM (RtP)   & 0.367 (0.377) & 0.393 (0.413) & 0.429 (0.459) & 0.459 (0.493) & 0.514 (0.575) \\
    PI-FGSMPP (RtP)  & 0.369 (0.379) & 0.450 (0.483) & 0.509 (0.555) & 0.558 (0.629) & 0.584 (0.657) \\
    NI-FGSM (RtP)   & 0.350 (0.359) & 0.355 (0.359) & 0.378 (0.389) & 0.447 (0.483) & 0.513 (0.559) \\
    SINI-FGSM (RtP)  & 0.345 (0.349) & 0.355 (0.361) & 0.377 (0.389) & 0.470 (0.509) & 0.552 (0.609) \\
    VMI-FGSM (RtP)   & 0.384 (0.401) & 0.464 (0.505) & 0.536 (0.589) & 0.565 (0.625) & 0.563 (0.625) \\
    \textbf{Ours}  & \textbf{0.407 (0.431)} & \textbf{0.505 (0.553)} & \textbf{0.584 (0.651)} & \textbf{0.594 (0.665)} & \textbf{0.600 (0.667)} \\
    \bottomrule
    \multicolumn{6}{l}{*The best scheme is highlighted by \textbf{bold}.}
    \end{tabular}
    }
    \label{tab:various_epsilon_VLM}
\end{table}

\paragraph*{Various values of $\epsilon$}
Table \ref{tab:various_epsilon_VLM} displays the average and majority voting scores of various adversarial attack methods, with $\epsilon$ ranging from $4/255$ to $64/255$. Consistent with the findings in \cref{subsec:quantitative_results_algorithm}, our proposed method outperforms the baseline methods across all $\epsilon$ values. Notably, despite the latent-based I-FGSM method having more degrees of freedom, our \texttt{Replace-then-Perturb} method surpasses it even at smaller $\epsilon$ levels. At higher $\epsilon$ values, the \texttt{Replace-then-Perturb} method outperforms the latent-based adversarial attacks, and our \texttt{Contrastive-Adv} method further enhances the evaluation scores.

\begin{table}
    \centering
    \caption{Ablation study based on the average rank and Elo rating for the positive and negative questions. }
    \adjustbox{width=1\linewidth}{
    \begin{tabular}{c|cc|cc}
    \toprule
    \multirow{2}{*}{Methods} & \multicolumn{2}{c|}{Positive Questions} & \multicolumn{2}{c}{Negative Questions}  \\
    & Avg. Rank. & Elo  & Avg. Rank. & Elo  \\ \midrule
    \makecell{I-FGSM \\ + Latent-based} & 1.354 & 986 & 1.544 & 978\\\midrule
    \makecell{I-FGSM \\ + Replace-then-Perturb} & \textbf{1.241} & \textbf{1008} & 1.495 & 990\\\midrule
    \makecell{Conrstive-Adv \\ +Replace-then-Perturb} & 1.287 & 1005 & \textbf{1.282} & \textbf{1031}\\
    \bottomrule
    \end{tabular}
    }
    \label{tab:elo_pos_neg}
\end{table}

\subsection{Positive Questions}

Unlike negative questions, where adversarial perturbations alter the answers, positive questions require answers to remain consistent after perturbation. 
Table \ref{tab:elo_pos_neg} reveals that cosine similarity scores are nearly identical across all methods for positive questions. 
Consequently, we ranked three methods---1)Latent-based I-FGSM, 2) Replace-then-Perturb-based I-FGSM, and 3) Replace-then-Perturb-based \texttt{Contrastive-Adv}---according to cosine similarity. 
The results demonstrate that both proposed methods outperform the baseline methods, with \texttt{Replace-then-Perturb} slightly surpassing Latent-based I-FGSM and \texttt{Contrastive-Adv} exhibiting marginally lower performance due to its triplet loss function. Additionally, when compared alongside negative question results, our methods consistently show significant improvements over the baselines.

\begin{figure*}
    \includegraphics[width=1\linewidth]{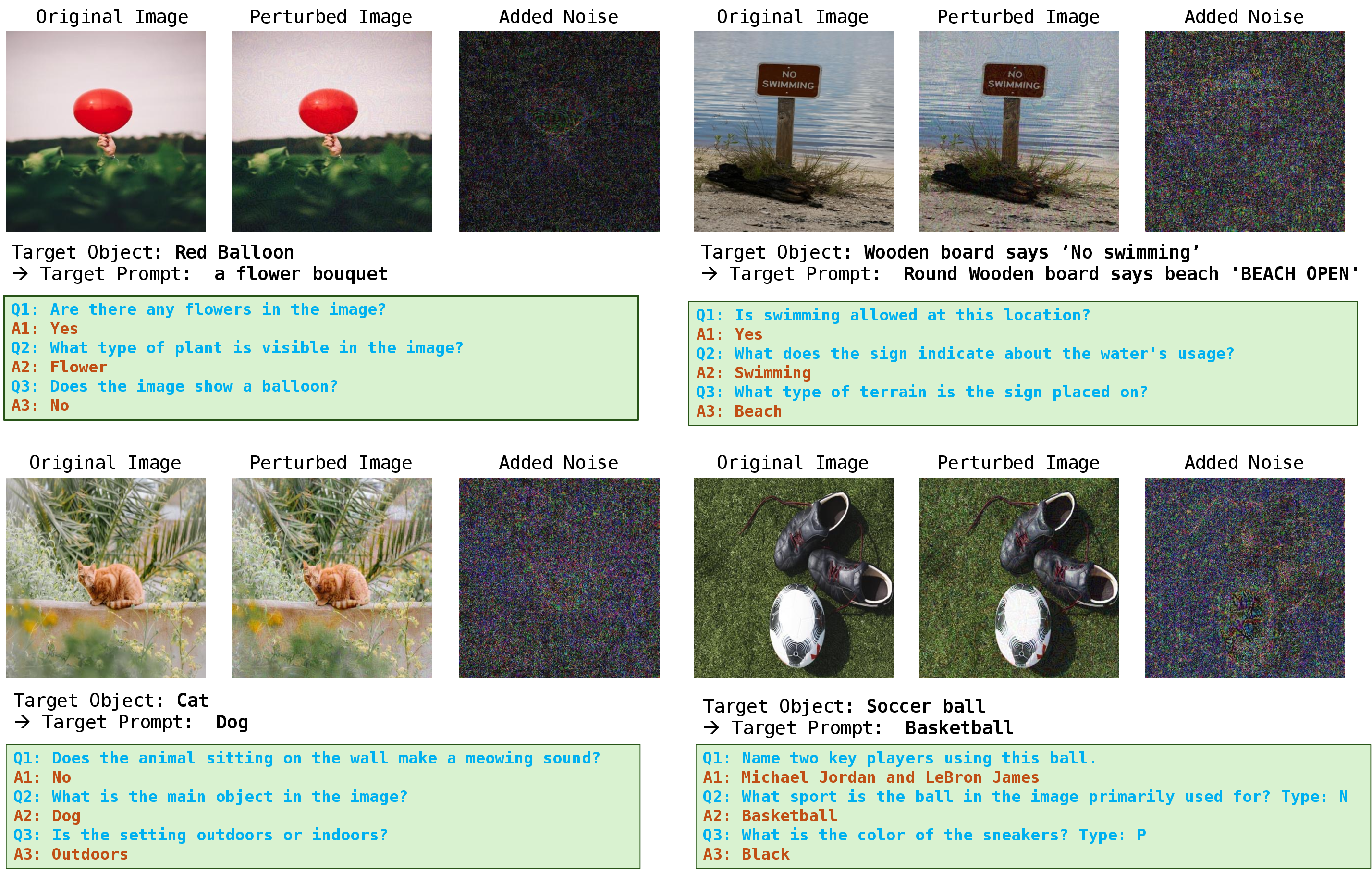}
    \caption{Illustrations of the adversarial examples generated by our proposed method (\texttt{Replace-then-Perturb} and \texttt{Contrastive-Adv}, where $\epsilon=16/255$ and $T=200$. In these illustrations, we depict the original image, perturbed image, and added noise, where the added noise images are  \textbf{10 times} amplified for visualization. }
    \label{fig:graphical_2}
\end{figure*}

\begin{table*}[!ht]
    \caption{Experimental results of the proposed method and baselines in VLM-based metrics, where $\epsilon=16.0/255$ and $T=200$, where the target VLM is LLAVA 1.6.  }
    \centering
    \adjustbox{width=1\linewidth}{
    \begin{tabular}{c|c|c|cccccccc}
    \toprule
    \multirow{2}{*}{Evaluation VLM} & \multirow{2}{*}{Metrics} & \multirow{2}{*}{Latent-based~\cite{zhou2023advclip}}   &  \multicolumn{7}{c}{Replace-then-Perturb \textbf{(Ours)}} \\
    & & & I-FGSM & PI-FGSM & PI-FGSMPP & NI-FGSM & SINI-FGSM & VMI-FGSM & Contrastive-Adv \textbf{(Ours)} & \makecell{Contrastive-Adv \textbf{(Ours)} \\ + VMI-FGSM} \\
    \midrule
\multirow{8}{*}{Majority vote} & BLUE & 0.439  & 0.489  & 0.411  & 0.541  & 0.373  & 0.377  & 0.581  & 0.561  & \textbf{0.619}  \\
 & GLUE & 0.435  & 0.479  & 0.395  & 0.543  & 0.343  & 0.353  & 0.585  & 0.571  & \textbf{0.651}  \\
 & ALL miniLM & 0.313  & 0.373  & 0.275  & 0.443  & 0.210  & 0.244  & 0.499  & 0.497  &\textbf{0.579}  \\
 & BGE-M3-COLBERT & 0.325  & 0.369  & 0.269  & 0.465  & 0.208  & 0.224  & 0.529  & 0.537  & \textbf{0.629}  \\
 & BGE-M3-DENSE & 0.317  & 0.381  & 0.273  & 0.469  & 0.214  & 0.238  & 0.523  & 0.529  & \textbf{0.581}  \\
 & BERT & 0.357  & 0.417  & 0.329  & 0.495  & 0.275  & 0.297  & 0.539  & 0.537  & \textbf{0.605}  \\
 & CLIP (image-txt) & 0.389  & 0.447  & 0.387  & 0.497  & 0.325  & 0.347  & 0.541  & 0.547  & \textbf{0.579}  \\
 & CLIP & 0.355  & 0.411  & 0.339  & 0.495  & 0.263  & 0.287  & 0.541  & 0.559  & \textbf{0.613}  \\
\multirow{8}{*}{avg} & BLUE & 0.417  & 0.450  & 0.407  & 0.489  & 0.378  & 0.384  & 0.508  & 0.501  & \textbf{0.535}  \\\midrule
 & GLUE & 0.409  & 0.441  & 0.391  & 0.484  & 0.359  & 0.365  & 0.508  & 0.502  & \textbf{0.549}  \\
 & ALL miniLM & 0.330  & 0.377  & 0.312  & 0.432  & 0.265  & 0.283  & 0.478  & 0.480  & \textbf{0.537}  \\
 & BGE-M3-COLBERT & 0.347  & 0.386  & 0.313  & 0.456  & 0.268  & 0.279  & 0.503  & 0.505  & \textbf{0.564}  \\
 & BGE-M3-DENSE & 0.337  & 0.386  & 0.310  & 0.452  & 0.270  & 0.282  & 0.498  & 0.497  & \textbf{0.542}  \\
 & BERT & 0.375  & 0.415  & 0.355  & 0.461  & 0.316  & 0.328  & 0.501  & 0.493  & \textbf{0.541}  \\
 & CLIP (image-txt) & 0.382  & 0.421  & 0.380  & 0.460  & 0.335  & 0.347  & 0.490  & 0.495  & \textbf{0.522}  \\
 & CLIP & 0.359  & 0.407  & 0.351  & 0.468  & 0.294  & 0.310  & 0.502  & 0.515  & \textbf{0.557}  \\
    \bottomrule
    \multicolumn{10}{l}{*The best scheme is highlighted by \textbf{bold}.}
    \end{tabular}
    }
    \label{tab:LLAVA_1_6_results}
\end{table*}









\paragraph*{Summary}

In this paper, we propose two methods for generating adversarial examples for VLMs: \texttt{Replace-then-Perturb} and \texttt{Contrastive-Adv}. We address the challenge of generating appropriate target feature vectors for replacing target objects with target prompts, a limitation in existing methods. The \texttt{Replace-then-Perturb} method leverages zero-shot semantic segmentation and inpainting to create target feature vectors. Building upon this, the \texttt{Contrastive-Adv} method further enhances the performance of adversarial examples. To evaluate our methods, we introduce a dataset comprising images, target objects, target prompts, and positive/negative questions.

\paragraph*{Limitations}

Our work has two primary limitations. First, we do not evaluate the effectiveness of our methods against existing defense mechanisms against adversarial examples~\cite{shah2019cycle,yang2021defending}, as this was outside the scope of our study. Second, our focus is on VLMs with separate text and visual encoders. Further studies are needed to assess the applicability of our methods to recent VLM architectures, such as \cite{team2024chameleon}.

\appendices

\appendices

\section{Additional Experimental Results}

\subsection{Additional Graphical Results}

Figure \ref{fig:graphical_2} illustrates additional adversarial examples generated by our proposed method. In the first example, adversarial noise successfully alters the phrase from ``red balloon'' to ``a flower bouquet,'' leading the target VLM to recognize the latter instead of the former. Similarly, the second to fourth examples demonstrate that our method effectively manipulates the target VLMs to perceive the target objects as the specified prompts through adversarial perturbations.

\subsection{LLAVA 1.6 Results}

Table \ref{tab:LLAVA_1_6_results} presents the quantitative results for LLAVA 1.6 as the target neural network model. Consistent with the findings in Tables \ref{tab:algorithm-based-metrics} and \ref{tab:VLM-based-metrics}, our \texttt{Contrastive-Adv} method outperforms most baseline adversarial attacks except for VMI-FGSM. Although VMI-FGSM achieves slightly higher scores, it is approximately ten times more computationally intensive due to its variance computation step. To further validate our approach, we introduced an enhanced \texttt{Contrastive-Adv} variant, where the perturbation step in Lines 10-11 of Algorithm \ref{alg:contrastive-adv} is replaced with VMI-FGSM as defined in Equations \eqref{eq:vmi-fgsm-1}, \eqref{eq:vmi-fgsm-2}, and \eqref{eq:vmi-fgsm-3}. The results in Table \ref{tab:LLAVA_1_6_results} show that this modified method significantly outperforms all baseline methods.

\bibliographystyle{IEEEtran}
\bibliography{egbib}

\end{document}